# Autonomous damage assessment of structural columns using low-cost micro aerial vehicles and multi-view computer vision

Sina Tavasoli[1], Xiao Pan[2], T. Y. Yang[3], Saudah Gazi[4], Mohsen Azimi[1]

[1] Ph.D. candidate, Department of Civil Engineering, University of British Columbia, sina.tavassoli@gmail.com
[2] Research fellow, Department of Civil Engineering, University of British Columbia, xiao.pan@civil.ubc.ca
[3] Professor, Department of Civil Engineering, University of British Columbia, yang@civil.ubc.ca
[4] Undergraduate intern student, Department of Civil Engineering, University of British Columbia

## 1. ABSTRACT

Structural columns are the crucial load-carrying components of buildings and bridges. Early detection of column damage is important for the assessment of the residual performance and the prevention of system-level collapse. This research proposes an innovative end-to-end micro aerial vehicles (MAVs)-based approach to automatically scan and inspect columns. First, an MAV-based automatic image collection method is proposed. The MAV is programmed to sense the structural columns and their surrounding environment. During the navigation, the MAV first detects and approaches the structural columns. Then, it starts to collect image data at multiple viewpoints around every detected column. Second, the collected images will be used to assess the damage types and damage locations. Third, the damage state of the structural column will be determined by fusing the evaluation outcomes from multiple camera views. In this study, reinforced concrete (RC) columns are selected to demonstrate the effectiveness of the approach. Experimental results indicate that the proposed MAV-based inspection approach can effectively collect images from multiple viewing angles, and accurately assess critical RC column damages. The approach improves the level of autonomy during the inspection. In addition, the evaluation outcomes are more comprehensive than the existing 2D vision methods. The concept of the proposed inspection approach can be extended to other structural columns such as bridge piers.

## 2. INTRODUCTION

Past earthquakes showed that a lot of structures needed to be quickly evaluated to guarantee the functionality of society and to improve the resiliency of cities [1-6]. Traditional post-earthquake damage evaluations mostly rely on visual observation by humans. This is time-consuming, and the inspector's expertise has a considerable impact on the accuracy of the evaluations [7] [8].

Columns are seen as an important member in post-earthquake evaluations of buildings as they are the main load-carrying systems in structures [9]. After an earthquake, the column might undergo damage and lose lateral strength. This reduction of lateral capacity may cause axial failures linked to the degradation of vertical load capacity [10]. ATC-20 (ATC 1989) [9] and ATC-20-2 (ATC 1995) [11] give suggestions about the categorization of concrete frame structures, mostly based on observable column damage. ATC-20 (ATC 1989) [9] states that a detailed visual evaluation of columns is necessary to distinguish between the damage states (red, yellow, and green). While the techniques described in these documents do successfully reach the desired level of structural condition assessment, it is typically challenging to tag every building in an affected area as either safe or unsafe right after an earthquake. This is due to the shortage of experienced inspectors, the amount of labor necessary for each phase of assessment, and the vast number of inspections itself [9].

Recently, vision-based SHM has developed as a practical, effective, and affordable method for structural health monitoring [12-14]. Vision-based techniques generally require a straightforward camera setup and image-processing algorithms to assess the damage of interest. In recent years, it has been shown that vision-based techniques are effective for various SHM applications such as low-cost and economical structural vibration measurement [15-19], and damage detection of reinforced concrete (RC) structures [20-25], steel structures [26-29], masonry structures [30-32] and structural bolted assemblies [29] [33-36]. These studies were focused on developing image processing algorithms to analyze 2D images. A little or no attention was paid to automating the data collection process. Besides, in many existing studies, as the image was captured from a single camera view of a structural component, the evaluation result can be sensitive to the location where the image was captured.

In recent years, unmanned aerial vehicles (UAVs) have been utilized as an effective data-collecting method for different purposes, including reconnaissance, agriculture, facility inspection (such as power lines), rescue, and mapping. Within civil engineering, applications of UAVs have been investigated in recent years [37]. Typically, GPS is used to carry out the localization of UAVs in an outside setting where there are few barriers. In an interior setting, the building will often block GPS signals. As a result, the location computation is usually inaccurate. To overcome this issue, Simultaneous Localization and Mapping (SLAM) [38] is introduced to make UAVs independent of GPS. SALM requires advanced sensors, such as radar, sonar, and lidar, in order to gather information to pinpoint the UAV's location [5] [39] [40]. These sensors provide dense and accurate data. However, they have a large payload and high prices and often require a UAV of medium to large-size for carrying them. This might not be affordable to many researchers and engineers. Also, it is challenging or even impracticable for medium and large-size UAVs to maneuver in extremely small interior areas (such as short hallways and doors) or post-disaster sites after a major earthquake [44]. More recently, a research study employed a small-size UAV (namely nano aerial vehicle, or NAV) for indoor inspection of RC structures [41]. Despite the accuracy being satisfactory, the NAV used in this study has only one distance sensor, which may limit its obstacle avoidance capability in a highly complex environment. Therefore, there is a need to design a drone of a relatively small size at a relatively low cost, while still carrying enough sensors to achieve some level of autonomy in navigation, obstacle avoidance and inspection in confined interior areas. For this purpose, micro aerial vehicles (MAVs) with various appropriate sensors can be designed.

To address these limitations, a novel semi-autonomous MAV-based data collection and damage evaluation pipeline is proposed to detect and assess column damages at high accuracy and low cost. The proposed methodology comprises a semi-autonomous MAV-based data collection approach, vision-based column localization, and multi-view vision-based damage evaluation. Experimental studies have been conducted on RC columns. The results indicate that the proposed methodology can be used to effectively detect RC columns, collect images from multiple views, and accurately identify and localize critical RC column damages. Also, it has been shown that the presented methodology can reduce the level of expertise required for piloting the MAVs for inspection purposes in indoor environments. The proposed methodology enhances the level of autonomy during inspections and provides a more comprehensive inspection compared to existing studies that used a single view for damage detection.

## 3. METHODOLOGY

The aim of programming the MAV is to improve the level of autonomy during inspections. However, achieving a fully autonomous inspection remains very challenging and is an active research topic in the field. In this study, to limit the scope, the control of MAV is semi-automated as described below. *Figure 1* summarizes the steps of the methodology. First, the MAV is manually controlled by a pilot while streaming data in real time to the local PC (A). The live-streamed image data will be analyzed in real-time to detect potential columns in a frame (B). If any column is detected, the MAV switch from manual control to autopilot (C). Then, the autonomous data collection algorithm will be activated to collect image data from the detected column (D). Finally, after the completion of data collection, the MAV switches from autopilot to manual control (E).

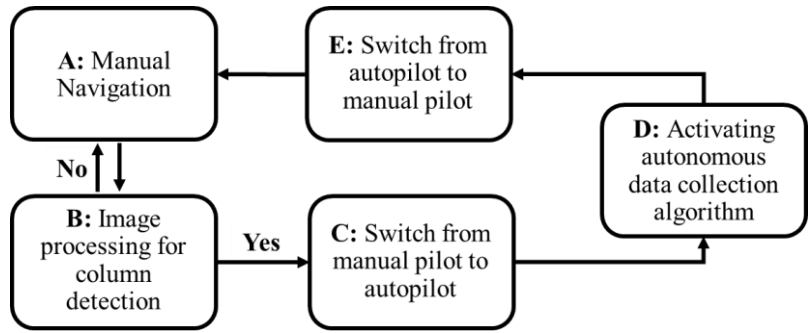

*Figure 1: flowchart summarizing semi-autonomous data collection procedures*

### 4. Autonomous data collection algorithm

The MAV is manually controlled and sends real-time images to the local PC where a vision-based algorithm is employed to find a column with a bounding box. Once a column is detected, the MAV switches from manual pilot to autopilot. The MAV is programmed to gradually approach the detected column to guarantee that the area of the detected bounding box exceeds 10% of the picture frame in order to retain appropriate resolution in the region of interest. The distance between the current position of the MAV and the column is determined using the LiDAR distance sensor installed on the MAV.

The MAV initiates a move on a circle with a radius equal to the distance between the column and itself. During the circular movement, the MAV stays focused on the detected column and captures photos on every predefined interval of rotational movement (e.g., 30 degrees). The captured images will be processed in real time for damage evaluation. In order to avoid striking potential obstacles during the rotation, the MAV will constantly receive feedback from side ultrasound distance sensors. In case, the MAV encounters an obstacle, it stops and reverses its circular motion around the column. If the MAV rotates for 360 degrees without encountering an obstacle, the algorithm assumes that a full scanning is completed and terminates the autonomous data collection in order to switch to manual control. In an indoor environment, there could be several scenarios for column locations. *Figure 2* demonstrates three possible scenarios of column and MAV positions for data collection in indoor areas. The columns might be in the center of a hall *Figure 2* (a), a corner of two walls *Figure 2* (b), and in a wall *Figure 2* (c). After the successful accomplishment of the autonomous data collection, the MAV switches from autopilot to manual control. *Figure 3* demonstrates a real-world example of an interior column and *Figure 4* demonstrates a 3D example of a whole scanning algorithm.

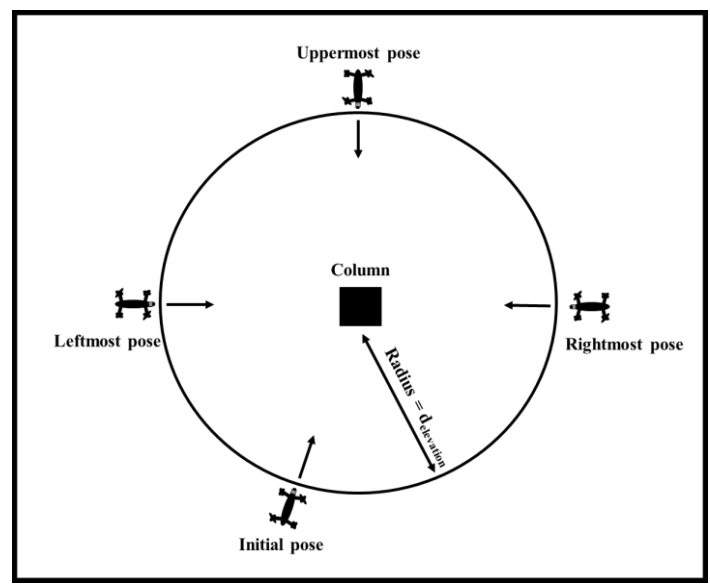

(a)

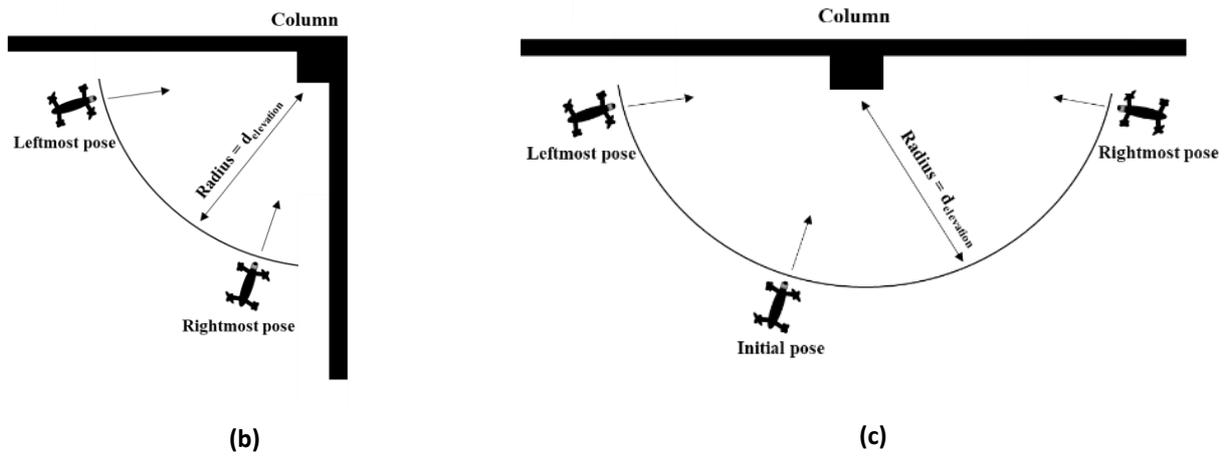

Figure 2: Three possible scenarios of column and MAV positions

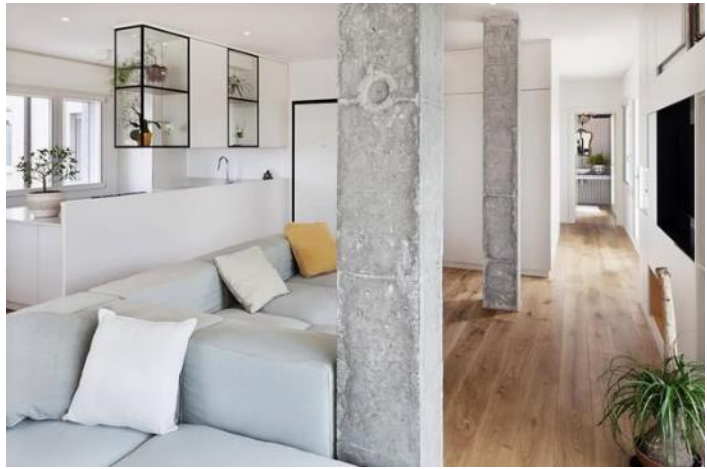

Figure 3: Real-world example of an interior column [42]

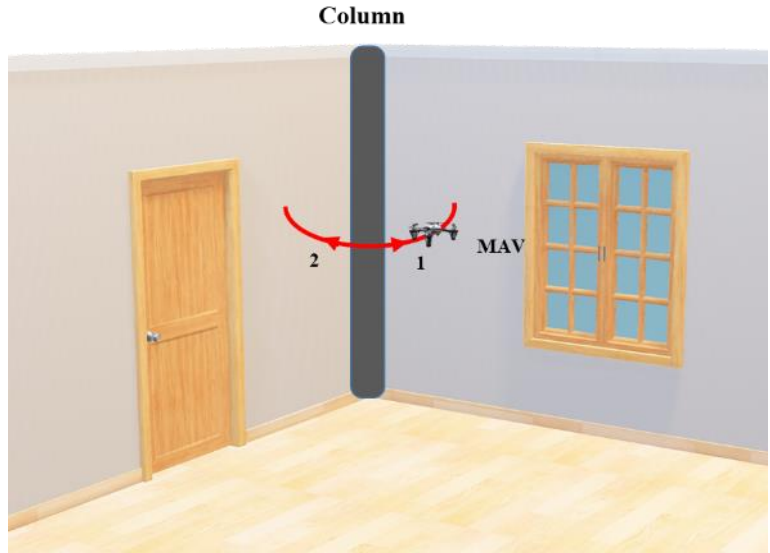

*Figure 4: 3D demonstration of the data collection process*

## 5. Vision-based damage detection

The damage types of concrete structures generally include concrete cracks, concrete spalling, concrete scaling, steel reinforcement exposure, steel reinforcement buckling or fracture, etc. Much existing literature has attempted to propose various methods for different types of damage. To limit the scope, in this study, two critical RC column damages are considered according to FEMA P58 [43], including the concrete spalling and exposed steel reinforcement, which contribute to the severe damage state of the RC structures. Other damage features such as cracks can be achieved by numerous existing methods. Within this context, two pretrained CNN-based vision methods were adopted from [41] to detect the two critical damage types.

As reviewed in Section 1, much of the existing vision-based SHM literature is focused on developing imaging processing techniques to analyze RGB images. Although these studies show promising accuracy in damage recognition and localization when processing a single RGB image, there lacks a further discussion on how the damage evaluation of results from individual images can be translated to a damage state of a structural component. More specifically, to illustrate such a limitation, *Figure 5* shows a concrete column example where images were captured from multiple views. If the computer vision algorithm is only applied to one of the cases shown in *Figure 5* (b) and *Figure 5* (c), the inspection outcome will lead to "light damage" and "no damage", respectively, which is different from the real damage state of the column, i.e., "severe damage". Therefore, it is evident that fusing the damage detection results from multiple image views will lead to a more comprehensive inspection of the column.

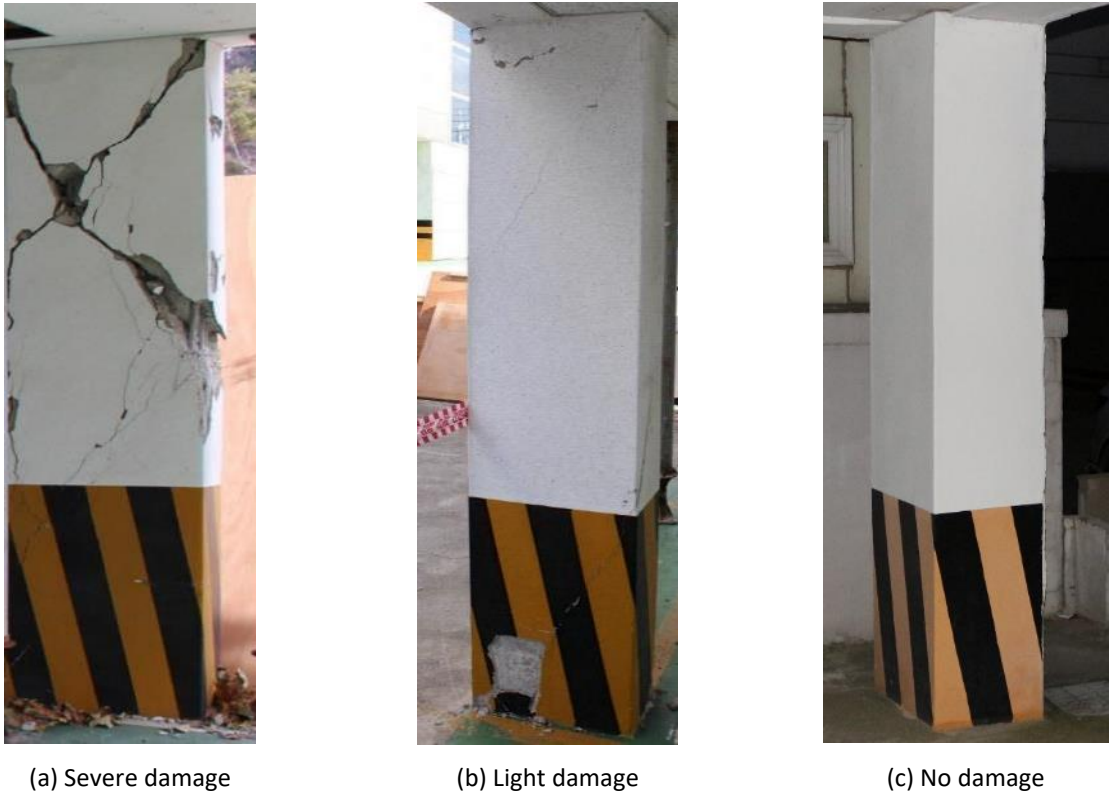

(a) Severe damage           (b) Light damage           (c) No damage

*Figure 5: Images captured from multiple views of a structural column with damaged and undamaged sides.*

In this study, one of the main goals is to address the above limitations. As shown in Section 2.1, the MAV has been programmed to automatically collect images from multiple views of the structural columns. The collected images will be processed by the pretrained YOLOv2 and DeepLabv3+ to recognize critical damages such as exposed steel reinforcement and concrete spalling. The damage diagnosis of a structural column is achieved by taking the worst-case scenario from the processing of images captured from multiple views.

## 6. IMPLEMENTATION

## 7. Description of the MAV and the collision avoidance module

In this study, a programmable MAV, Parrot Anafi, is utilized together with a number of additional sensors (*Figure 6*). Control commands will be sent to the MAV from a local PC. The MAV selected has a cutting-edge monocular camera that can record 4K video and 21MP HDR images. It possesses a 180° tilt camera gimbal that can be operated to change the camera pitch while in flight and a 3-axis sensor-driven picture stabilization. It also has an IMU and GPS, which are used to localize the MAV's position while it is autonomous data collection phase. The MAV is equipped with a PCB which hosts different sensors and an ESP32 module. In order to measure the distances between the MAV and the objects around, a 1D distance LiDAR sensor (TFs-mini) and three ultrasonic distance sensors are installed on the PCB and mounted on the MAV. The data from the distance sensors will be sent to the ESP32 module. The ESP32 uses a WI-FI interface to send the data back to the local PC. The power supply of the sensors and the ESP 32 is provided through two 5-volt rechargeable lithium-polymer batteries. *Figure 6* summarizes the distance sensors and the arrangement used in this paper.

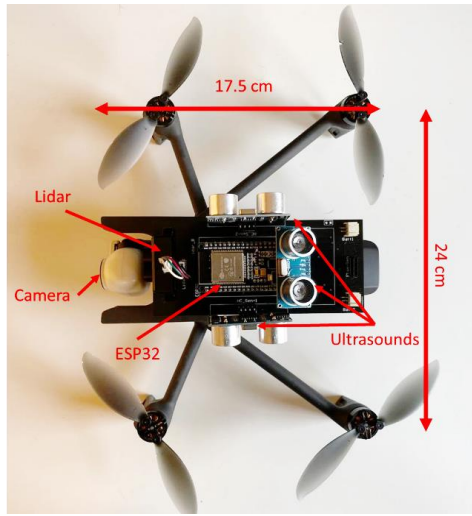

*Figure 6: The designed PCB for collision avoidance mounted on the MAV*

## 8.  Experiment and results

To evaluate the effectiveness of the proposed methodology, the algorithm is tested on damaged RC columns in the structural laboratory at The University of British Columbia. *Figure 7* demonstrates the setup of the experimental test. The algorithm is tested under three scenarios where the column is located 1- in a center without any obstacle around, 2- in a corner, and 3- in the middle of a wall span. During the experiment, the MAV was controlled manually, and it was streaming the image data back to the local PC in real time. Once a damaged RC component got detected, the MAV switched from manual pilot to autopilot and initiated the autonomous data collection algorithm described in Section "Autonomous data collection algorithm". *Figure 8* demonstrates the trajectory of the MAV and the location of the columns during the experiment.

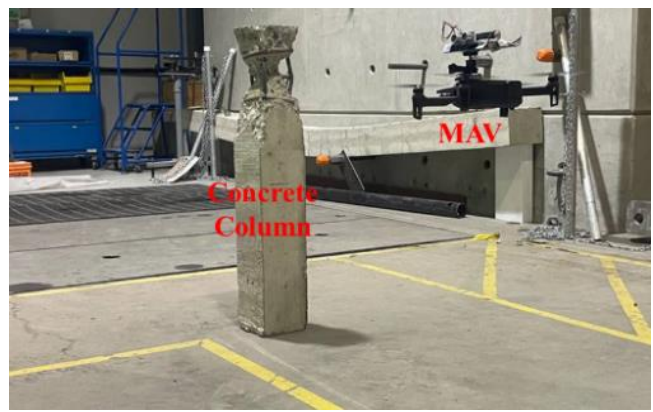

*Figure 7: Implementation setup*

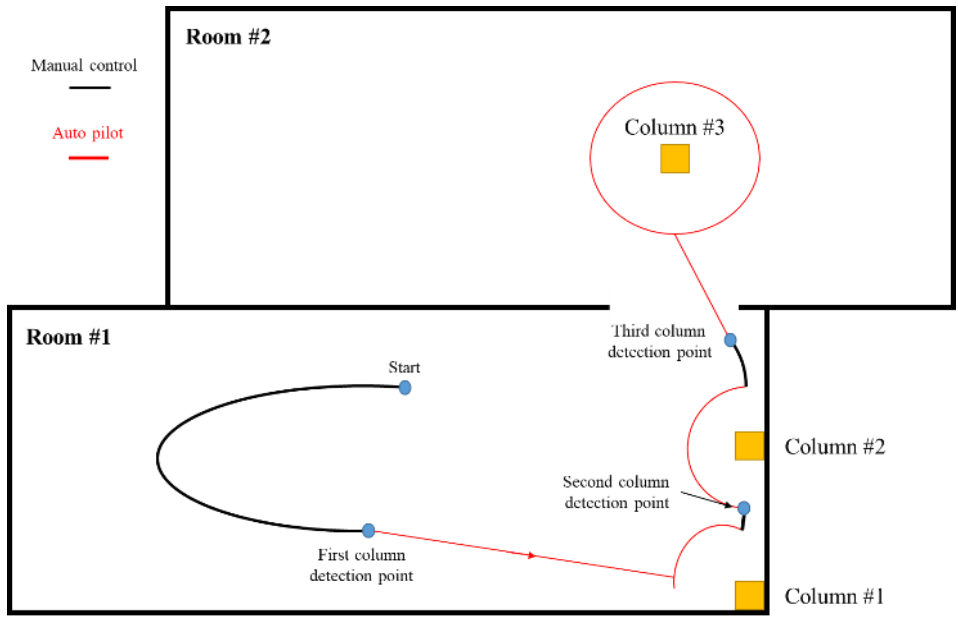

*Figure 8: Trajectory of the manual and autopilot navigation*

During the autonomous scanning, the photos are captured at every predefined interval (e.g., 30 degrees) and at both ends where MAV detects an obstacle on the sides. This results in having many images from the columns from different viewing angles. *Figure 9* to *Figure 11* demonstrate some of the images collected during the autonomous data collection from the damaged RC columns. The images captured during the data collection are processed in real time to detect concrete column damages including rebar exposure and spalling. During the autonomous data collection, the MAV successfully prevented obstacles and provided image data of different viewpoints from the RC columns. In addition, the vision-based damage detection methods accurately detected and localized the damage features in all three cases.

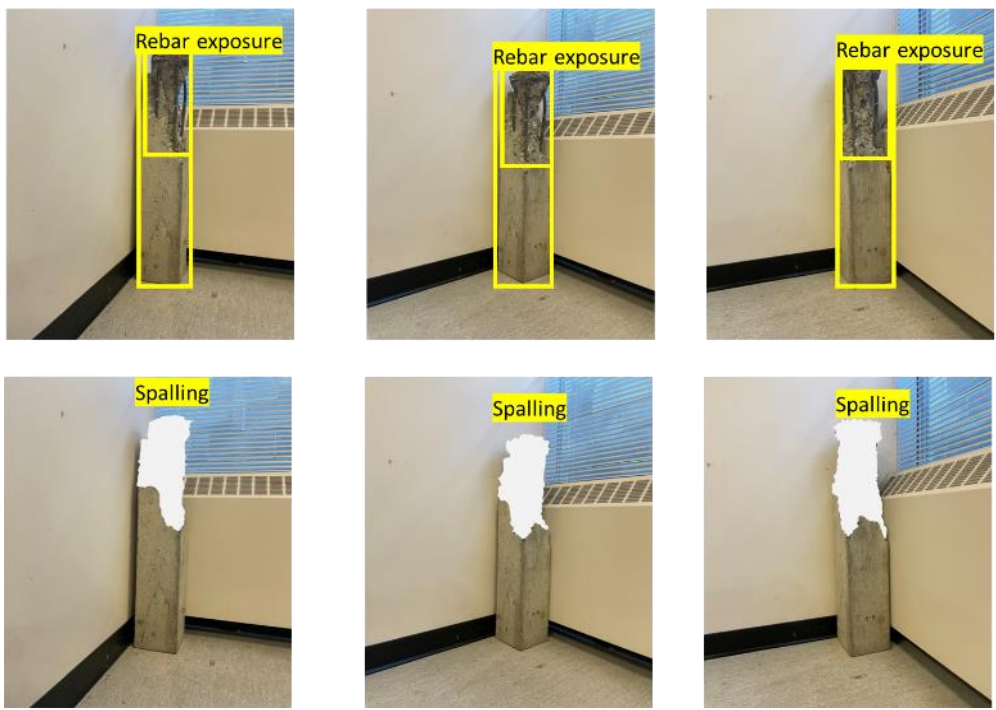

*Figure 9: rebar exposure and spalling captured by MAV in column #1*

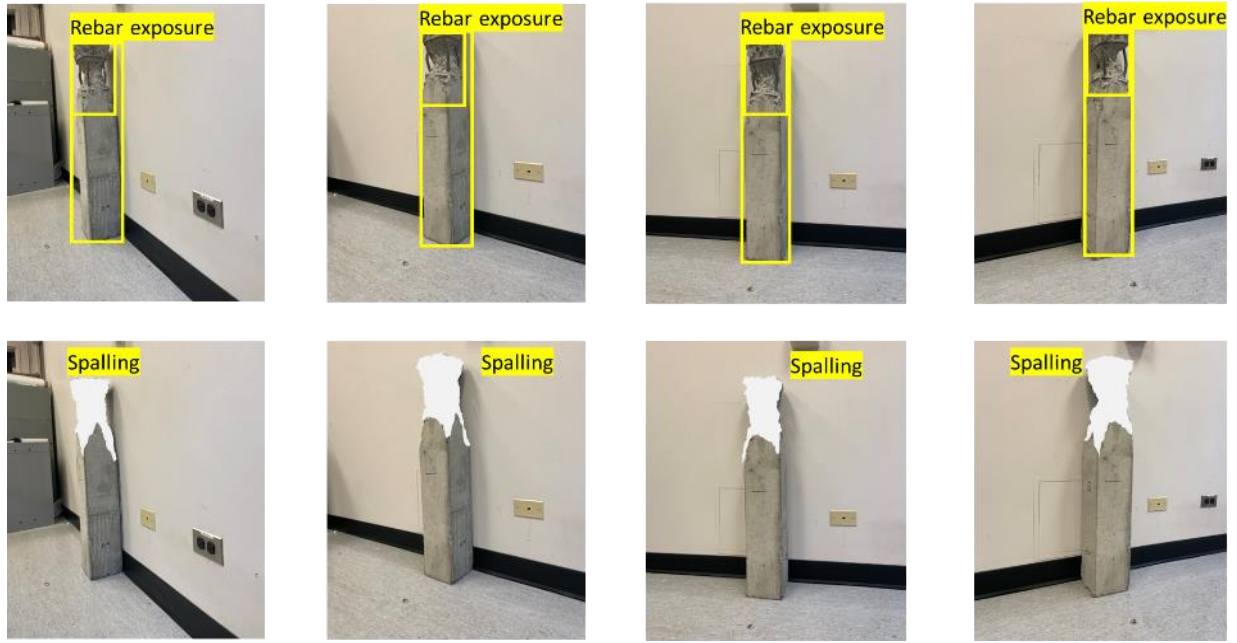

*Figure 10: rebar exposure and spalling captured by MAV in column #2*

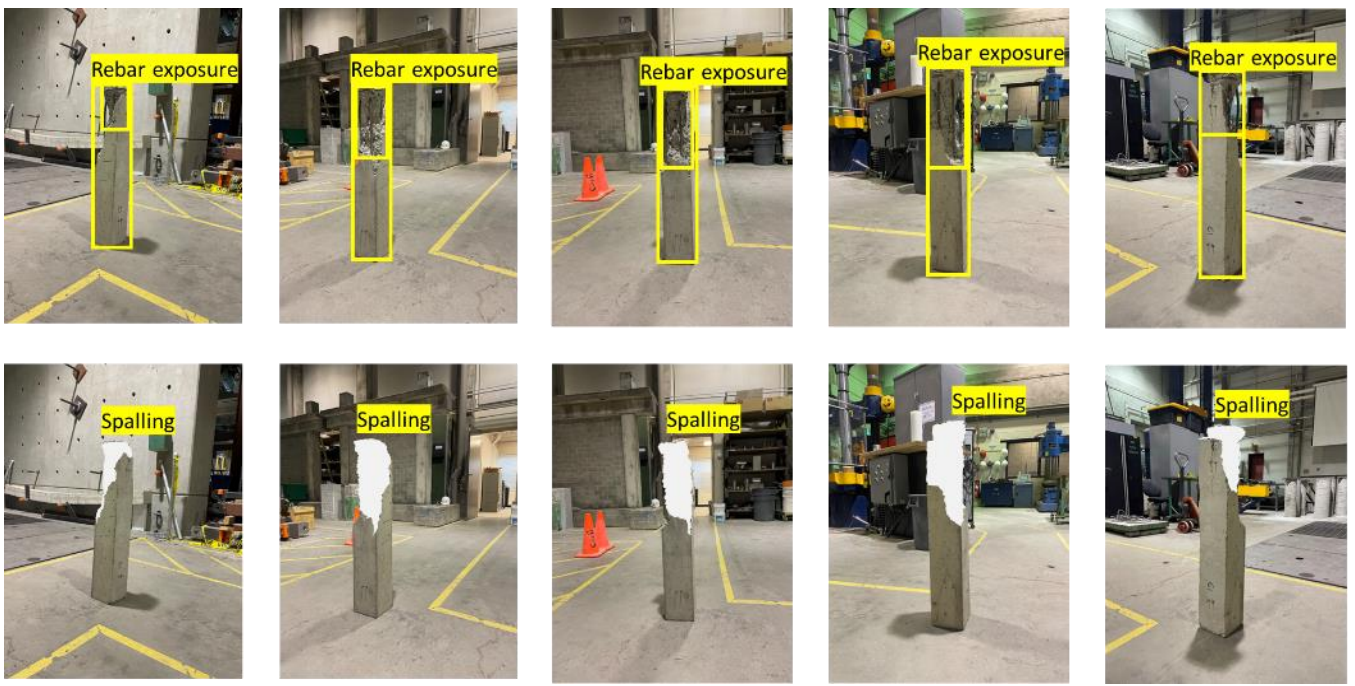

*Figure 11: rebar exposure and spalling captured by MAV in column #3*

## 9. CONCLUSION

In this research, a new semi-autonomous pipeline for indoor column data collection and damage inspection using low-cost MAVs is presented. A PCB hosting Lidar and three ultrasound sensors are designed to be integrated with MAV for autonomous data collection and obstacle avoidance. The integration of the designed PCB, sensors, and the MAV allows the MAV to detect obstacles in all directions. The presented methodology has been tested on three RC columns in an indoor scenario. The proposed autonomous data collection methodology can reduce the level of expertise required for piloting the MAVs for inspection purposes in indoor environments. The proposed vision-based damage detection methods can be used to detect multiple RC damage types such as concrete spalling and exposed steel reinforcement. Overall, the presented framework improves the level of autonomy and has the potential to accelerate the inspection process for full-scale buildings in the future.

## 10. LIMITATIONS AND FUTURE STUDY

Limitations in the present study are concluded from multiple perspectives:

As for the data collection process, the proposed method is not fully autonomous. The autonomous data collection procedures are developed only at the local level, i.e., to target a single component. While the proposed procedures enhance the level of autonomy, there is still a gap in achieving fully autonomous inspection using UAVs for multiple structural components, or a complex structural system. As a further study, the proposed method will be integrated with global path planning and navigation algorithm such as advanced SLAM. In that case, additional sensors such as stereo camera or 2D LiDAR will be installed on the present MAV.

As for the damage assessment, the proposed multi-view vision methods are targeted to multiple RGB images captured at multiple views of the structural component. While such methods provide a more comprehensive inspection compared to methods that rely on a single-view image, they still cannot offer a more detailed evaluation, such as localization and quantification of damage in 3D space. As an active ongoing study, the authors are exploring vision-based 3D reconstruction for structural components, where a series of images for a structural component will be collected to reconstruct the dense point cloud of the component in 3D space. Advanced 3D point cloud processing methods will be further developed to evaluate the structural damage in 3D space. Further, the 3D reconstruction procedures will be integrated with more advanced autonomous navigation and data collection procedures to achieve a fully autonomous and even more comprehensive evaluation by the MAV.

Lastly, as for the capability of the drone, the proposed MAV has its advantages and flexibility in navigating in a broader vertical space compared to unmanned aerial vehicles (UGV). However, the MAV-based inspection can be hampered by the closure of inter-room gates. In such cases, a collaborative scheme between the MAV and UGV should be developed where the UGV can be deployed to open the gate for the MAV.